\documentclass{article}

\PassOptionsToPackage{numbers, compress}{natbib}

\title{Satellite Imagery and AI: A New Era in Ocean Conservation, from Research to Deployment and Impact (Version. 2.0)}

\usepackage[final]{compsust_2023}

\usepackage[utf8]{inputenc} 
\usepackage[T1]{fontenc}    
\usepackage{hyperref}       
\usepackage{url}            
\usepackage{booktabs}       
\usepackage{amsfonts}       
\usepackage{nicefrac}       
\usepackage{microtype}      
\usepackage{xcolor}         
\usepackage{graphicx}
\usepackage{amsmath}

\author{%
  Patrick Beukema* \\
   \And
  Favyen Bastani \\
  \And
  Yawen Zhang \\
    \AND
  Piper Wolters \\
  \And
  Henry Herzog 
  \And
  Joe Ferdinando 
  \AND
  Allen Institute for AI (Ai2) \\
  *correspondence: \texttt{patrickb@allenai.org}
}

\begin{document}

\maketitle

\begin{abstract}
Illegal, unreported, and unregulated (IUU) fishing poses a global threat to ocean habitats. Publicly available satellite data offered by NASA, the European Space Agency (ESA), and the U.S. Geological Survey (USGS), provide an opportunity to actively monitor this activity. Effectively leveraging satellite data for maritime conservation requires highly reliable machine learning models operating globally with minimal latency. This paper introduces four specialized computer vision models designed for a variety of sensors including Sentinel-1 (synthetic aperture radar), Sentinel-2 (optical imagery), Landsat 8-9 (optical imagery), and Suomi-NPP/NOAA-20/NOAA-21 (nighttime lights). It also presents best practices for developing and deploying global-scale real-time satellite based computer vision. All of the models are open sourced under permissive licenses. These models have all been deployed in Skylight, a real-time maritime monitoring platform, which is provided at no cost to users worldwide.\footnote{An earlier version of this manuscript won best paper at the 2023 NeurIPS Computational Sustainability workshop.} 

\end{abstract}

\section{Introduction}

Unprecedented environmental catastrophes compounded by ruthlessly efficient fishing are pushing our oceans to the brink. Entire species have gone missing seemingly overnight. Following the disappearance of billions of snow crabs from the Bering Sea in 2022, the fishery closed for the first time in its history \cite{snowcrab}. Worldwide, it is estimated that 34\% of fisheries are unsustainably harvested \citep{FAO2020}, a concerning trend that continues to escalate.

Remote sensing data coupled with artificial intelligence provides a means to monitor and detect global maritime activity including threats to marine ecosystems. No single satellite can provide adequate coverage of the entire planet. However, leveraging many satellites with a variety of passive and active sensors enhances the likelihood of identifying destructive behavior as it occurs enabling successful interventions. Large, publicly available image data from a diverse constellation of satellites enables real-time monitoring of the entirety of the world's oceans.

This paper provides an overview of four recently developed computer vision models designed for global-scale near real-time vessel detection. In aggregate, these models process over 1.9 TB of satellite imagery per day, covering 100\% of all Exclusive Economic Zones (EEZs). Each of these models has been deployed in Skylight \cite{skylight}, a maritime intelligence platform, supporting international conservation efforts through real-time monitoring. Skylight is provided for free to users worldwide, spanning 308 organizations and over 60 countries. 

\begin{figure}
  \centering
  \centerline{\includegraphics[scale=.33]{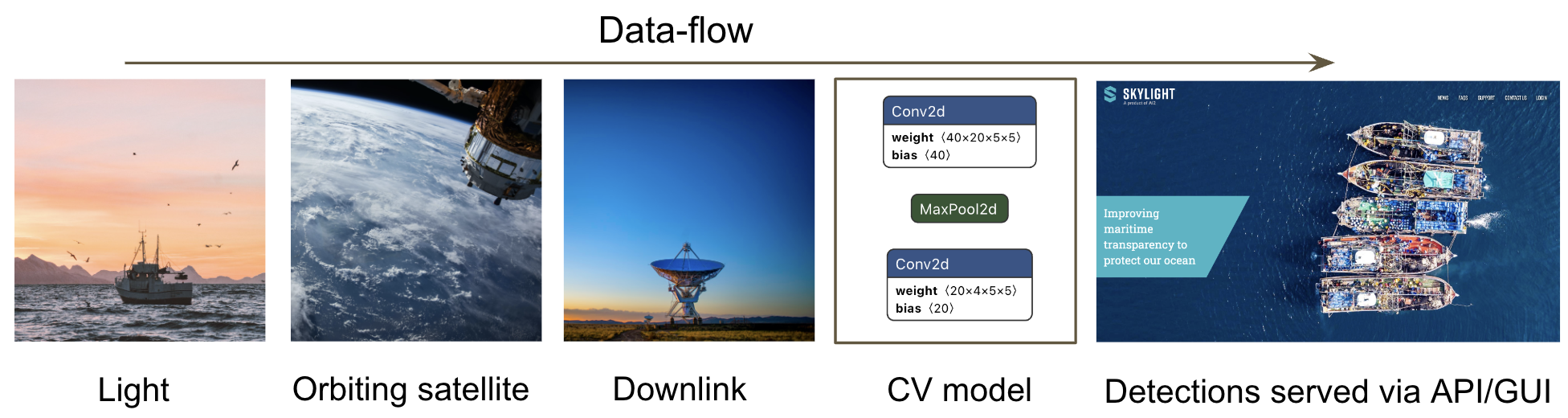}}
  \caption{Data-flow depiction of a real-time streaming computer vision service for vessel detection in satellite imagery. An orbiting satellite images a vessel. The image is downlinked to a ground station, copied to Skylight owned servers and processed by a computer vision model \cite{netron}. The resulting vessel detection is reported to our users through a GUI and available via an API.}
\label{dfd}
\end{figure}

\section{Building computer vision models for maritime intelligence}

 \begin{table}
  \caption{Overview of each satellite model/API in Skylight}
  \label{overview}
  \centering
  \begin{tabular}{cccccccc}
    \toprule
    \cmidrule(r){1-2}
         & Provider  & \# sat   & PX (m) & Signal  & Revisit  & Latency   & Count (wk 4/24/25) \\ 
    \midrule
    VIIRS &  NASA  & 3 & 750  & watts  & 12 hrs &   1.5 hrs  & 95,764 \\
    Sentinel-1     &   ESA & 1 & 10     & radar  & 14 days &   5 hrs & 54,713 \\
    Sentinel-2     &   ESA   & 2 & 10   & optical & 5 days &  5 hrs  & 62,438  \\
    Landsat    & USGS   & 2 & 15-30   & optical & 8 days &  4 hrs & 32,214 \\
    \bottomrule
  \end{tabular}
\end{table}

Achieving high performance is critical for our users, who cannot afford to expend limited resources, such as fuel, on pursuing non-existent vessels.. However, while performance is paramount, there are other important considerations that anchor research and development, including minimizing latency and ensuring adequate interpretability. It is essential to report a vessel's presence as quickly as possible. Although the overall latency--from the vessel to the satellite, then to the model, API, and finally the user--is dominated by the downlink latency (see table \ref{overview} "Latency"), computational efficiency is important. This efficiency facilitates high throughput iteration and regular upgrading. In addition, the model outputs should be interpretable. If the model commits egregious errors or its reasoning is opaque, our users cannot (and should not) trust its outputs. For this reason, the platform outputs a simple crop centered on each vessel detection (Fig. \ref{fig:example_detections} D-F) to allow users to visually inspect every detection. We aim for transparency and share documentation about the model creation and ML strategy to help establish confidence in the machine intelligence.

In the following sections we provide a brief description of the unique characteristics and modeling strategy for each satellite. Table \ref{overview} provides a high level overview of each vessel detection service. The code and model architectures alongside complete processing pipelines and additional details have been open sourced on GitHub \cite{vvd, svd, sentinel2, landsat}.

\subsection{Vessel detection in VIIRS imagery}
\begin{figure}
  \centering
  \centerline{\includegraphics[scale=.13]{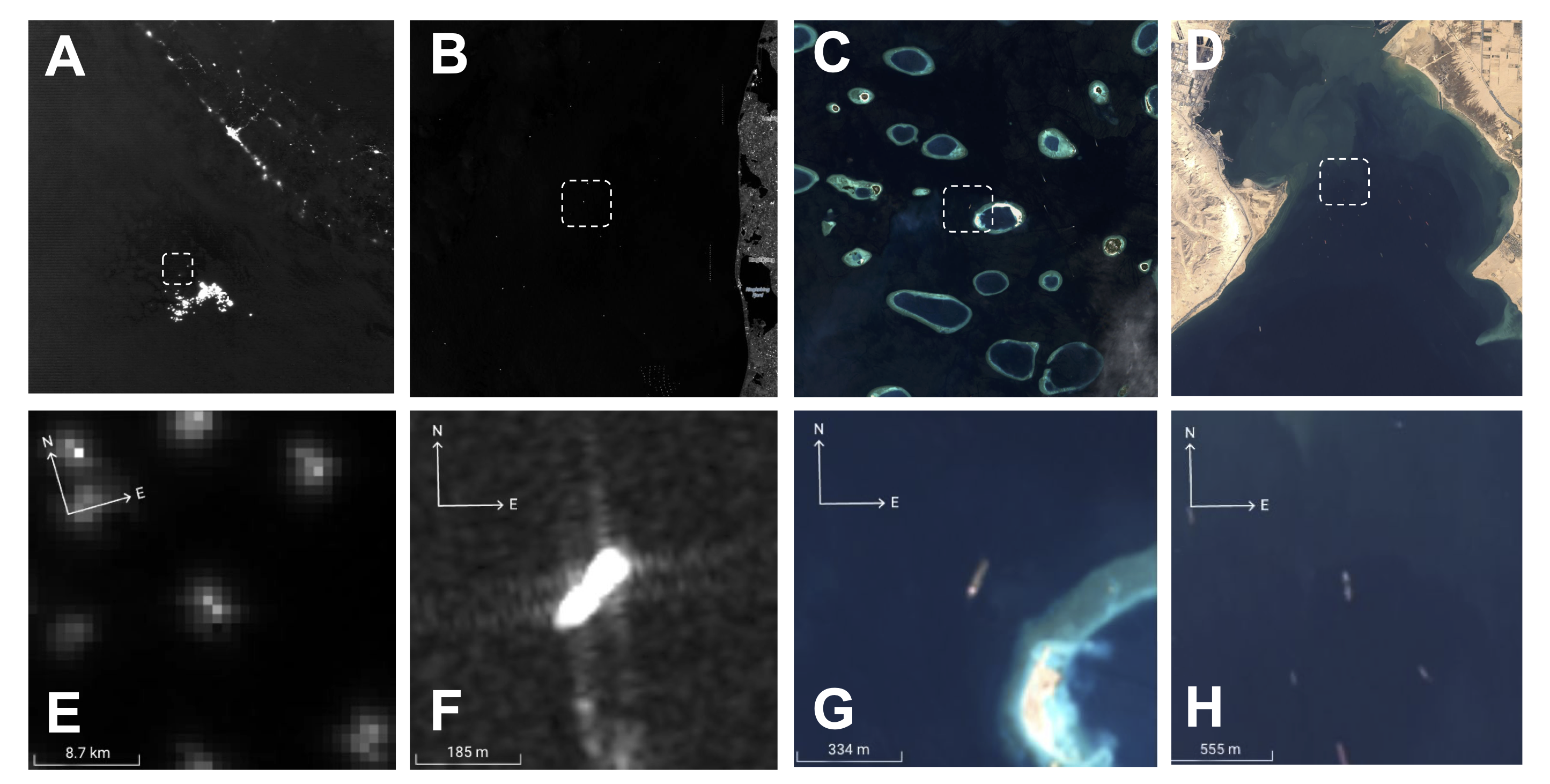}}
  \caption{Example satellite imagery (top row) and sample detections (bottom row) from a VIIRS image (A, E) near the Ecuadorian coast, an S1 image (B, F) from the North Sea, an S2 image (C, G) from the Maldives, and a Landsat image (D, H) from the Suez Canal. Scale bars are approximate. Confidence scores $>$ 0.95.}
  \label{fig:example_detections}
\end{figure}

The Visible Infrared Imaging Radiometer Suite (VIIRS) sensor on board the Suomi-NPP and NOAA-20 satellites collect visible and infrared images during both the day and night \cite{viirselv}. While not originally intended as a real-time vessel monitoring data source, the low latency ($\sim 2$ hrs), global coverage, satellite redundancy, and unique signal characteristics make the VIIRS sensor a useful tool in the fight against illegal fishing. However, the low spatial resolution (750m) precludes distinguishing vessels from non vessels prima facie (see example detection in Fig. \ref{fig:example_detections}E). Therefore, care must be taken to achieve high precision. 

The modeling strategy adopted a three stage approach. The first stage consisted of a classical computer vision model, trained without supervision, to extract all possible sources of light.  This was achieved with a simple 2D kernel. In the second stage, all known non-vessel light sources (lightning, gas flares, moonlit clouds, the northern and southern lights, and ionospheric particles from within the South Atlantic Anomaly) are removed through a series of postprocessing steps \cite{saa}. These non-vessel light sources often exhibit stereotyped distributional patterns (unlike vessels) that is amenable to rules based logic. Additionally, we implemented statistical tests to identify unusually geographically distributed vessels coincident with scan lines and suppressed false positives at the frame's extremeties due to the ``noise smile" \cite{Elvidge2015AutomaticBI} to control the false positive rate. The final stage involved filtering all positive detections through a regularly updated 2d CNN. This CNN was trained on human annotated image labels (correct/not) with four channels (nanowatts, land water masks \cite{viirs-land-masks}, moonlight, and clouds \cite{ackerman2017viirs, ackerman2017viirs_noaa20}). This model was specifically designed to run in resource constrained environments, requiring only modest hardware (2 GB RAM, no GPU). 

End-to-end deep learning based approaches were also evaluated. However, given the simplicity and limited spatial extent of the objects (1-2 pixels, as shown in \ref{fig:example_detections}E), and their sparse distribution, end-to-end deep learning based models required significantly more computational power to achieve performance comparable to our hybrid design. Our hybrid approach is designed to be highly efficient, which allows for regular and economical retraining using new labeled data. Such efficiency is particularly beneficial for continuoul improvement, ML specific continuous integration and continuous delivery (CI/CD) pipelines and model-specific unit/integration/regression testing. Thorough testing (especially within the CICD framework) is beneficial for preventing regressions during phases of fast-paced development, and is typically prohibitively expensive with conventional large-scale DNNs.

\subsection{Vessel detection in S1 imagery}
Sentinel-1 (S1) is a satellite constellation from the ESA providing 10 m/pixel synthetic aperture radar (SAR) imagery, which measures properties of energy reflections from the planet surface \cite{torres2012gmes, fletcher2012sentinel}. It captures Earth's land surface and coastal waters roughly every two weeks. The visual features are more complex than VIIRS, requiring detailed visual discrimination to distinguish vessels from islands and fixed marine infrastructure, and we found that deep learning methods were required. We created a dataset of 55,499 vessels (point labels) annotated by maritime experts. 

We developed a detection model consisting of a standard Faster-RCNN~\cite{fasterrcnn} head, with a customized backbone. The backbone consists of a small 13-layer fully-convolutional encoder, which outputs feature maps at four resolutions that are processed through a feature pyramid network (FPN)~\cite{fpn}. We adapt the backbone to input not only the current target S1 image (with two bands, VV and VH polarizations) in which vessels should be detected, but also one or more aligned historical images of the same region at different times. These historical images enable the model to learn that marine objects consistently in the same location, such as fixed platforms and islands, are unlikely to represent transient objects like vessels. We process the images independently through the backbone, and at each resolution we concatenate the features of the target image with the pooled features of the historical images and pass the result to the FPN. Note that an early version of this model architecture scored fourth place in the xView3 competition \cite{xview3}. The training data used here does not overlap with the xView3 data, and in contrast to xView3, is 100\% human annotated (rather than by machine).

\subsection{Vessel detection in S2 imagery}
Sentinel-2 (S2) is another ESA satellite constellation, providing optical imagery with four bands at 10 m/pixel, six bands at 20 m/pixel, and three bands at 60 m/pixel \cite{esa2012sentinel2}. It captures Earth's land surface and coastal waters roughly every 5 days. The same maritime experts annotated 43,102 vessels (point labels) in S2 imagery.  

The various S2 bands provide rich information about the physical objects present in a scene. For example, RGB bands already enable distinguishing most vessels from other marine objects, but additional bands can be leveraged to further improve accuracy due to the different reflectance signatures of vessels and other objects. Thus, we developed a detection model that, like our S1 model, uses a Faster-RCNN detection head, but we couple it with a much larger Swin Transformer~\cite{swintransformer} backbone that has sufficient parameters to perform complex analysis of the S2 bands.

We found that pre-training the backbone on SatlasPretrain~\cite{bastani2023satlaspretrain}, a large-scale remote sensing dataset, further improved performance. Unlike with S1, we did not observe a performance increase from inputting historical images. We speculate that this is because the visual signatures of vessels in S2 optical images are sufficiently distinct from stationary objects (e.g., platforms or wind turbines), so historical images are not needed. 

\subsection{Vessel detection in Landsat imagery}

Landsat 8 and Landsat 9 are part of NASA and the U.S. Geological Survey’s (USGS) long-running Landsat program~\cite{usgs2020landsat}, which has been providing continuous Earth observation data since 1972. The Operational Land Imager (OLI) onboard captures high-resolution imagery in nine spectral bands, ranging from visiable to shortwave infrared wavelengths. Spatially, these satellites provide imagery with a resolution of 30 m across most spectral bands, a higher 15-m resolution for the panchromatic band. Temporally, each satellite operates on a 16-day revisit cycle, and when combined, Landsat 8 and 9 deliver an effective revisit time of approximately 8 days. This frequent imaging capability allows for timely monitoring of physical objects.

In our initial experiments, we found that using a single detector with Landsat 8/9 resulted in many false positives that are caused by whitecaps, ice, clouds, and islands. This is likely due to Landsat 8/9’s lower resolution compared to Sentinel-2. To overcome this, we developed a two-stage detection model for Landsat 8/9 imagery. The first stage employs a detector with a Swin-v2-Base~\cite{swintransformer} encoder and Faster R-CNN head to identify potential vessel locations using bands B2 to B8 (covering RGB bands and shortwave infrared wavelengths). The detector is trained on 18,509 vessel labels from 7,954 patches, and it uses overlapping sliding windows and non-maximum suppression to manage large scenes and reduce duplicate detections. To enhance accuracy of the detections, the second stage uses a Swin-v2-Base classifier that examines the patches around the detected vessels to confirm their presence. This classifier is trained on approximately 2,000 expert-annotated detections, enabling it to effectively distinguish true vessels from false positives. Together, the two-stage pipeline significantly improves detection accuracy compared to using the detector alone. 

\subsection{Model evaluation, validation, and deployment}
While there is no single metric that we use to determine when a model is ready to be released to our users, each model is continuously evaluated against a variety of criteria as it passes from research, to staging, to deployment. During the research phase, the primary method of evaluation consists of offline F1 scores against large and randomly held out validation datasets (where possible). For S1, we also compared the model against a previous version that had been submitted to the xView3 competition (the new model improved from 70.1\% to 82.7\% F1). The S2 model was not part of an external competition, but exhibited a similar F1 score of 0.81. For Landsat, the detection model achieved an F1 score of 76.0\% and the classification model achieved an F1 score of 72.2\%. There was no large annotated dataset for the VIIRS model, however, we did compare its performance against the industry standard model on previously human validated frames \cite{Elvidge2015AutomaticBI}. Due to the fact that offline performance of VIIRS was limited, we supplemented this evaluation with extensive unit and integration tests (CI/CD), covering a variety of known failure modes (aurora, moonlit clouds, imaging artifacts, etc). 

Once all failure modes have been addressed, we transition models into an online staging environment that replicates production dataflows (i.e. streaming and real-time inference). Importantly, all of the inferenced data in the staging environment is new (out-of-sample) data. Models are evaluated in this staging environment for extended periods (months) by subject matter experts versed in maritime intelligence and regularly updated in response to that feedback.

Once confident that a new model is performant, we deploy models from staging into the production environment. In production, the primary method of evaluation consists of user feedback (internal and external) which gives us prompt notice of performance degradation or model drift. 

After deployment, models are regularly upgraded (monthly release cadence) in response to feedback and/or new information sources. It is worth noting that while the data sources are (largely) stable, ocean activity is highly dynamic. For example, marine infrastructure (wind turbines, oil platforms, etc) is constantly under construction. These dynamics must be tracked and addressed through regular maintenance to maintain high performance. For example, to improve precision of each of the above models, we recently added an additional postprocessing step that geofences false positive detections coincident with recently detected \cite{satlas-geospatial-2023} marine infrastructure produced by \citet{bastani2023satlaspretrain}. 

\subsection{Known Limitations}
As a general rule, for optical and radar-based imagery (Sentinel-1, Sentinel-2, and Landsat), model performance tends to decrease proportionally with the size of the object being detected. Typical sources of false positives for these models include whitecaps, small icebergs, and rock islands. A notable performance drop-off occurs around the pixel resolution (see table \ref{overview} for each sensor's resolution). However, it is possible to detect objects smaller than a pixel dimension because the models also leverage other contextual features, such as vessel wakes, which can be significantly larger than the vessels themselves. In practice, these models (excluding VIIRS) are able to accurately resolve vessels across a wide dynamic range, from vessels with widths from 1 -- 60 meters and from 3 -- 400+ meters in length. 

The VIIRS model, which relies on detecting light emissions, avoids many of these sources of false positives, because as such objects do not emit light. However, because the VIIRS sensor was originally designed to detect moonlit clouds (not vessels), it exhibits unique failure modes. Some false positives due to clouds still occur, particularly under the full moon where the light is bright and under adverse condtions reflects off of clouds producing vessel like features. We limit these by omitting detections underneath clouds (via a separate cloud detector), which reduces recall but ensures high precision.

All models prioritize precision over recall where trade-offs are necessary. This prioritization reflects user feedback, as high precision is critical to operational use cases. Users accept some reduction in recall to maintain consistently high precision, given that the models generally exhibit high recall. To improve precision, we also suppress detections coincident with marine infrastructure~\cite{marine_infra}. We also employ a high-resolution (10m) coastal point dataset ~\cite{lighthouse} which further improves precision by eliminating false positives due to land such as small rocky islands. This coastal data also improves recall by enabling recognition of vessels close to shore, on inland lakes, etc, which is not possible with coarse resolution coastal datasets.  

\subsection{Correlating satellite detections and GPS}
\begin{figure}
  \centering
  \centerline{\includegraphics[scale=.3]{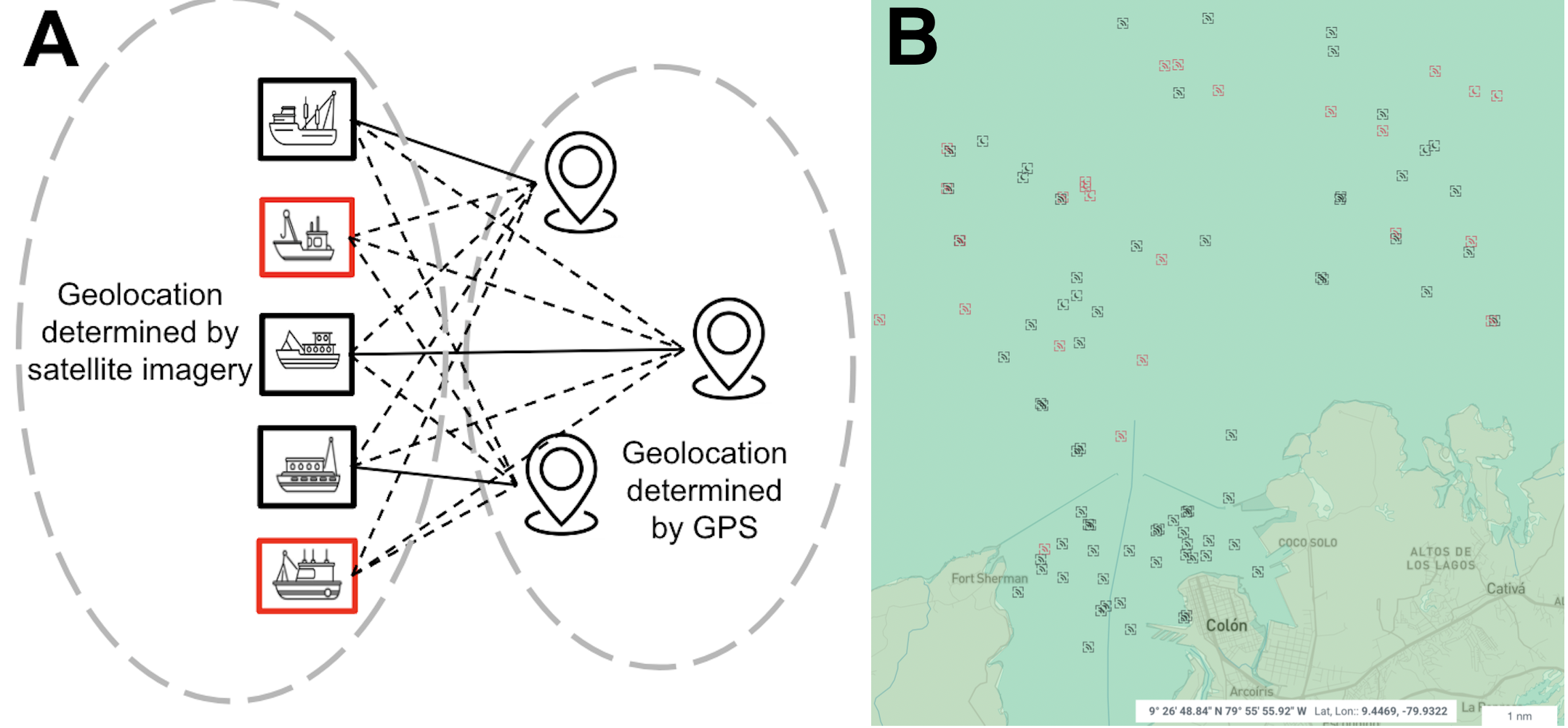}}
  \caption{A. Depiction of the correlation process. We compute the haversine distance between vessels in imagery and as located by AIS, then minimize the distance over the pairs. B. Panel from the Skylight UI showing radar, optical, and night lights detections near Col\'{o}n, (Panama Canal). Correlated = black, uncorrelated = red. }
  \label{fig:cor}
\end{figure}

Every vessel detection is augmented by the addition of GPS information, when available, provided by the Automatic Identification System (AIS). Most vessels broadcast their locations, but some do not. Notably it is possible to obfuscate one's position by suppressing AIS but quite challenging to evade detection from a satellite. Vessels that neglect to broadcast their locations, but are still visible under radar or satellite imagery are especially relevant to analysts.  In a typical satellite image, there are many detected vessels, and many possible candidate matches from AIS signals in the vicinity. Therefore, it is necessary to correlate the signals from these two information sources. Fig. \ref{fig:cor} shows a depiction of the correlation process. Matching geolocations from an image and as determined by GPS can be formulated as a simple minimum weight matching problem in bipartite graphs. We apply a Jonker-Volgenant algorithm \cite{jonker} as implemented by \citet{scikit} to assign matches.

\section{Best practices for shipping computer vision for maritime intelligence}
Deploying computer vision models into a real-time streaming context has offered many valuable lessons, particularly around closing the gap between offline batch and online streaming performance. 
\begin{itemize}
\item{Near real-time satellite data may exhibit unique features or artifacts that will be missed if training on historical imagery. If streaming/online inference is the goal, ensure that the model is trained against the same data that will ultimately be inferenced in production.}
\item{The performance of static models, i.e. those with frozen weights, is at risk of regression due to model and/or data drift. Ensure that model iteration is as simple and as automated as possible in order to facilitate seamless retraining from user feedback.}
\item{Allow ample time to empirically assess the model performance in real world conditions and at appropriate temporal and spatial scales to identify and correct problems as they occur. For example, VIIRS is highly sensitive to the lunar cycle.}
\item{Users cannot rely on  machine intelligence unless it is consistently available and reliable. Expect to dedicate significant engineering resources beyond modeling to ensure that the model is always online and maintained.}
\item{Best practices from software engineering, such as continuous integration and continuous deployment, unit and integration tests, static type checking, code quality enforcement, and documentation all accelerate research and development.} 
\item{Satellite imagery exhibits massive variation at global scale and therefore it can be challenging to anticipate how performance will degrade on out-of-distribution data. For example, after deployment we discovered hitherto unknown sources of false positives due to the Aurora Borealis/Australis for VIIRS, newly constructed wind turbines in the North Sea for the S1 model, and sargassum patches \cite{sargassum} in the Caribbean for the S2 model.}
\end{itemize}

\section{Equity considerations}
Real time vessel detections from each these models are provided through Skylight, a free maritime intelligence platform. The purposes of these models specifically, and Skylight more broadly, is to help nations protect their marine resources and promote ocean health. While we believe that the benefits of this technology (both within our platform and as open source repositories) outweigh potential risks, there is a possibility these models could be used for nefarious ends. We do not have a straightforward response that obviates these concerns and we do not take the decision to open source lightly. We chose to open source these models because we believe that both the machine learning research community and the conservation community using the vessel detections should have complete transparency and full access into the underlying models and logic. In addition, these models are made possible by the existing open geospatial community, as well as the data providers (NASA, ESA, USGS) which enables both historical analysis and real-time inference.

\begin{ack}
\subsection*{Allen Institute for AI (Ai2)}
All authors are employees at Ai2. Skylight is a product of Ai2. 
\subsection*{Computer Vision Annotation team} 
Ebenezer Aidoo (Ghana Navy), James Curtis Carter (Ghana Navy), Paige Roberts (Skylight)

\subsection*{Defense Innovation Unit}
The Defense Innovation Unit contributed funding which supported creation of the VIIRS, S2 and Landsat vessel detection models, and supported the improvements of the S1 model. The S1 model described in this paper was an extension of a previous model created at Ai2 team that was submitted to the \href{https://iuu.xview.us/}{xView3} competition. 

\subsection*{NASA \& LANCE (VIIRS)}
We acknowledge the use of data and imagery from NASA's Land, Atmosphere Near real-time Capability for EOS (LANCE) system (\url{https://earthdata.nasa.gov/lance}), part of NASA's Earth Observing System Data and Information System (EOSDIS). The data products that were used for VIIRS model are described here: \url{https://lance.modaps.eosdis.nasa.gov/viirs/}. The following products are used. For SuomiNPP, \texttt{VNP02DNB\_NRT} (light), \texttt{VNP03DNB\_NRT} (supporting data), \texttt{VNP02MOD\_NRT} (gas flares). For NOAA20, \texttt{VJ102DNB\_NRT} (light), \texttt{VJ103DNB\_NRT} (supporting data), \texttt{VJ102MOD\_NRT} (gas flares). In addition, we use the cloud masks created by the University of Wisconsin SSEC (\url{https://www.earthdata.nasa.gov/learn/find-data/near-real-time/viirs-a}) \cite{ackerman2017viirs_noaa20, ackerman2017viirs}. Additional details and code to use these data can be found on the VIIRS GitHub repository \cite{vvd}.

\subsection*{ESA (European Space Agency) and the Copernicus Data Space Ecosystem (S1 and S2)}
Data for the S1 and S2 constellations are available at \url{https://dataspace.copernicus.eu/}. Additional details of the specific products used for each satellite model are provided below. 

\begin{itemize}
\item{S1: Level-1 GRD (Ground Range Detected) data were used. Models were trained with both VV and VH polarization modes.} 
\item{S2: Level-1C data (orthorectified top-of-atmosphere reflectance, with sub-pixel multispectral registration). In addition to the L1C data, we also apply cloud detection to suppress false positives due to clouds. To do so we use the s2 cloud detector (s2cloudless) available from PyPI \cite{sentinel-2cloud-detector}. More details on this algorithm are available in \cite{s2cloudless}.}  
\end{itemize}

Note that we previously used Copernicus Open Access Hub which is deprecated as of October 2023, and replaced by the Copernicus Data Space Ecosystem. 

\subsection*{USGS (Landsat 8-9)}

We acknowledge the use of Landsat 8-9 imagery courtesy of the U.S. Geological Survey \url{https://www.usgs.gov/centers/eros/science}. The Landsat 8-9 Collection 2 Level-1 scenes data was obtained via the USGS M2M API client. For more details and code, see the Landsat GitHub repository \cite{landsat}.

\end{ack}

\bibliography{references}

\end{document}